# Exploring Emoji Usage and Prediction Through a Temporal Variation Lens

**Francesco Barbieri**[♣] **Luís Marujo**[♡] **Pradeep Karuturi**[♡]
**William Brendel**[♡] **Horacio Saggion**[♣]
[♣] Large Scale Text Understanding Systems Lab, TALN, UPF, Barcelona, Spain
[♡] Snap Inc. Research, Venice, California, USA

[♣]{name.surname}@upf.edu, [♡]{name.surname}@snap.com

## Abstract

The frequent use of Emojis on social media platforms has created a new form of multimodal social interaction. Developing methods for the study and representation of emoji semantics helps to improve future multimodal communication systems. In this paper we explore the usage and semantics of emojis over time. We compare emoji embeddings trained on a corpus of different seasons and show that some emojis are used differently depending on the time of the year. Moreover, we propose a method to take into account the time information for emoji prediction systems, outperforming state-of-the-art systems. We show that, using the time information, the accuracy of some emojis can be significantly improved.

## 1 Introduction

Emojis are frequently used on social media (Snapchat, Twitter, Facebook, Instagram) and on communication platforms (Whatsapp, Messenger). In turn, they create a new form of multimodal communication, wherein images are used to enrich standard text messages. Over the past few years, the interest in emoji research has increased with several studies which contributed to emoji semantics [BRS16, ERA+16, WBSD17a, WBSD17b, BCC18], sentiment analysis [NSSM15, HGS+17, KK17, RPG+18], automatic emoji prediction [BBS17, FMS+17] and multimodal systems [CMS15, CSG+18, BBRS18]. However, to the best of our knowledge, the temporal dimension of emojis has not been addressed in past research. In this paper we explore the temporal correlation between emoji usage and events during the year, and we show that temporal information helps disambiguate emoji meanings. For example, the four leaf clover emoji 🍀 is usually associated with good luck wishes, while in March, the same emoji indicates parties and drinking, due to St. Patrick day. In addition, some emojis are naturally associated with specific seasons. (e.g.,🎅 during Christmas and 🏖️ in Summer), or specific hours (e.g.,🌃 by night, and 🌄 in the morning). We show that considering temporal information helps predict emojis, including those that are not time-specific such as 💪 and ❤️.

Table 1: Most Frequent Emojis over different Seasons.

| | | | | | | | | | | | | | | | |
|---|---|---|---|---|---|---|---|---|---|---|---|---|---|---|---|
| **Spring** | ❤️ | 😍 | 😂 | 🔥 | 💕 | ☀️ | 😎 | 🙌🏼 | 💯 | 💙 | ✨ | 😊 | 🎉 | 😘 | 💪 |
| **Summer** | ❤️ | 😍 | 😂 | 🔥 | ☀️ | 😎 | 💕 | 💙 | ✨ | 🙌🏼 | 🇺🇸 | 💯 | 🎉 | 💪 | 😊 |
| **Autumn** | ❤️ | 😂 | 😍 | 🔥 | 💕 | 💯 | ✨ | 🙌🏼 | 💙 | 🎄 | 😊 | 🎉 | 🎃 | 😎 | 😘 |
| **Winter** | ❤️ | 😂 | 😍 | 🔥 | 🎄 | ❄️ | 💯 | 💕 | 😊 | 🙌🏼 | 😘 | 😎 | 🎉 | ✨ | 💪 |

## 2 Datasets

We first collected a corpus $\mathcal{C}^{🐦}_{us}$ of more than 100 million English tweets, posted only in the U.S.[1] from October 2015 to November 2017, and retrieved via the Twitter API[2]. We then extracted two datasets out of $\mathcal{C}^{🐦}_{us}$.

### 2.1 Seasonal Emoji Dataset

We divide our initial corpus into four subsets (tweets posted in Spring, Summer, Autumn and Winter) to In: S. Wijeratne, E. Kiciman, H. Saggion, A. Sheth (eds.): Proceedings of the 1st International Workshop on Emoji Understanding and Applications in Social Media (Emoji2018), Stanford, CA, USA, 25-JUN-2018, published at http://ceur-ws.org

[1] To remove spatial and cultural influence on data [BKRS16].
[2] https://dev.twitter.com/streaming/overview

Table 2: Nearest-Neighbour (NN) results. Season/pair combinations, and all indicates the number of emojis that are common in all the season NN.

| Emoji | 🎶 | 🎤 | 🎼 | 🍦 | 🎸 | 🌙 | 🎵 | 🐠 | 🐟 | 🍭 | 😂 | 💖 | 😏 | 💓 | 🍰 |
|---|---|---|---|---|---|---|---|---|---|---|---|---|---|---|---|
| **Spr−Sum** | 9 | 9 | 9 | 9 | 9 | 9 | 10 | 10 | 9 | 10 | 8 | 8 | 8 | 8 | 8 |
| **Spr−Aut** | 10 | 10 | 10 | 9 | 9 | 9 | 10 | 9 | 9 | 10 | 8 | 8 | 8 | 8 | 10 |
| **Spr−Win** | 9 | 9 | 9 | 10 | 9 | 9 | 9 | 9 | 9 | 9 | 8 | 9 | 8 | 9 | 10 |
| **Sum−Aut** | 9 | 9 | 9 | 10 | 9 | 10 | 10 | 9 | 9 | 10 | 9 | 8 | 8 | 9 | 8 |
| **Sum−Win** | 10 | 9 | 10 | 9 | 9 | 10 | 9 | 9 | 9 | 9 | 8 | 8 | 8 | 8 | 8 |
| **Aut−Win** | 9 | 9 | 9 | 9 | 9 | 10 | 9 | 9 | 9 | 9 | 8 | 9 | 9 | 9 | 10 |
| **All** | 9 | 9 | 9 | 9 | 9 | 9 | 9 | 9 | 9 | 9 | 8 | 8 | 8 | 8 | 8 |
| **Emoji** | 🚀 | 💣 | 🍀 | 🏀 | 🚨 | 🏆 | ⚡ | ❗ | ⭐ | 🎓 | ➡️ | 🌲 | 🎨 | 🌈 | 📍 |
| **Spr−Sum** | 5 | 5 | 3 | 6 | 2 | 6 | 6 | 2 | 7 | 2 | 6 | 8 | 3 | 2 | 3 |
| **Spr−Aut** | 3 | 4 | 4 | 5 | 3 | 6 | 4 | 5 | 4 | 3 | 5 | 3 | 3 | 1 | 1 |
| **Spr−Win** | 4 | 3 | 3 | 4 | 4 | 4 | 3 | 3 | 5 | 1 | 5 | 1 | 1 | 2 | 2 |
| **Sum−Aut** | 3 | 5 | 5 | 4 | 3 | 5 | 6 | 1 | 3 | 2 | 6 | 3 | 1 | 5 | 2 |
| **Sum−Win** | 4 | 4 | 8 | 2 | 4 | 4 | 4 | 2 | 5 | 5 | 5 | 1 | 2 | 2 | 0 |
| **Aut−Win** | 5 | 3 | 5 | 5 | 3 | 3 | 2 | 4 | 4 | 2 | 3 | 7 | 1 | 3 | 0 |
| **All** | 2 | 2 | 2 | 1 | 1 | 1 | 1 | 1 | 1 | 1 | 1 | 0 | 0 | 0 | 0 |

study the variation of emojis usage across different seasons (Section 3). Table 1 shows the 15 most frequent emojis of each season. We can see that while emojis including 😂, ❤️ and 😍 are always the most common, other emojis are select season-specific: 🎃 in Autumn, ❄️ and 🎄 in Winter, and ☀️ in Spring and Summer.

### 2.2 Large Scale Emoji Prediction Dataset

We retain from $\mathcal{C}^{🐦}_{us}$ tweets containing only one emoji, and only if that emoji belongs to the set of top 300 most frequently occurring emojis. The final dataset for emoji prediction is composed of 900,000 tweets, with 3,000 tweets per class. In previous work, we experimentally observed that using more than 3,000 tweets per class does not significantly improve the prediction accuracy.

## 3 Does the Emoji Semantic and Usage Change Over Seasons?

Emoji semantics are difficult to analyze due to the subjective nature of emojis meanings, especially when it comes to describing emotions. Nevertheless, we study emoji semantic by association, i.e. we describe an emoji with either a set of semantically close emojis, or by emoji pair co-occurrence in the same tweet. To this extent, we train skip-gram word embeddings models [MSC+13] on the four different subsets (Spring, Summer, Autumn and Winter) of our seasonal dataset. Each model embeds emojis within a high dimensional space (300 dimensions, 6 tokens window) where distance metrics translate to semantic closeness and co-occurrence. Following [BKRS16] we first evaluate emoji semantics by describing each emoji with its k-Nearest-Neighbours (k-NN) for each season. Secondly, for each model, we produce a correlation matrix that encodes the semantic correlation of pairs of emojis appearing in the same tweet. We then compare the four matrices to see if their correlation statistics is preserved across different seasons.

Table 3: The 10 NN emojis of the pine emoji 🌲 computed with respect to the semantic space of the four seasons. Emojis on the left are the closest ones.

| | | | | | | | | | | |
|---|---|---|---|---|---|---|---|---|---|---|
| **Spring** | 🌳 | 🍃 | 🍂 | 🌾 | 🌿 | 🍁 | 🌱 | 🗻 | 🌄 | ⛺ |
| **Summer** | 🌳 | 🌾 | 🌿 | 🍃 | ⛺ | 🌄 | 🍄 | 🌱 | 🌻 | 🗻 |
| **Autumn** | 🎄 | 🎅 | 🎁 | ☃️ | 🌳 | ❄️ | 🍃 | ❤️ | 🌿 | 💚 |
| **Winter** | 🎄 | ☃️ | 🎅 | 🎁 | ❄️ | ❤️ | 💚 | ⛺ | ✨ | ☺️ |

### 3.1 Analysis From Emoji k-NN Description

For each season, emojis are associated to their k-NN, with k=10. We then look at the overlap of these NN for each emoji among different models. This way, we can investigate if a specific emoji shares the same set of NN across distinct seasons and thus if that emoji preserves its meaning across seasons. We also measure the number of NN that overlap in all the seasons to find emojis with smaller meaning variation during the year. Results are shown in Table 2. In the top half of the table, we notice that emojis related to music, animal, sweets, and emotions are not influenced by seasonality, i.e. they have the same set of Nearest-Neighbours in the four season-specific vector models (10-NN overlap $\geq 8$).

The emojis for which their NN set varies the most across seasons are the ones listed in the bottom half of Table 2. Many are some sport-related emojis including 🏀 and 🏆 probably due to seasons of the year that are include more sports events that other periods. Also the emoji 🎓, used in a school/university graudation context, seems to change meaning across seasons. The Nearest-Neighbours of these emojis are party and heart-related emojis in Spring, while school-related emojis in Autumn.

Another season-dependant example is the pine emoji 🌲 (Table 3). The pine tree is associated with vegetation, camping and sunrise-related emoji in Spring and Summer, while in Autumn and Winter it co-occurs with Christmas-related emoji. The gift emoji 🎁 present a similar behaviour: in Spring and Summer the three nearest neighbors are 🎈, 🎊 and 🎂, while in Winter the three closest emojis are 🎅, 🎄, 🌲.

### 3.2 Analysis From Emoji Correlation

We evaluate how the semantics of pairs of emojis is preserved across different seasons. To this extent, we compute for each season a $300 \times 300$ correlation matrix, where the correlation of emoji $i$ and $j$ is encoded as the cosine similarity between their 300-D feature vectors extracted from the season model. We then compare pairs of seasons by evaluating the Pearson's correlation between their respective matrices. The most correlated matrices are Spring and Summer (0.871) while the lowest correlation is between Spring and Winter (0.837). However, all the matrices are highly correlated, suggesting that only a small subset of emojis have their semantic varying across seasons.

Table 4 shows for each pair of seasons the pairs of emojis with the highest difference in similarity across those seasons. The differences between Spring and Summer (first column) does not look as significant as the differences between other seasons. We can spot few interesting cases. For example the pair 🏆 💉 is more correlated during Autumn and Winter than during Spring and Summer. This is due to a famous case of doping in sport occurred during that Autumn/Winter of 2016. The pair 💨 🏆 characterizes track related competitions happening during Spring and Summer. Another interesting case is the gift emoji 🎁 that in Autumn and Winter relates to a Christmas gift, as it is highly correlated to ❄️ and 🌲, while in Spring and Summer it is mostly used as a birthday gift as it is associated to emojis like 🍰 and 🎈. The case of the pair 🔫 🎓 can relate to either mass-shooting events or could simply suggests that in Autumn students have a hard time with the beginning of the new school year. One of the emojis that seems to be used differently in Autumn and Winter is the Skull 💀, likely due to the usage of this emoji during Halloween time.

# 4 How Does Temporal Information Help Emoji Prediction?

In this section we evaluate how temporal information can improve the accuracy of emoji prediction models. We use the same experimental settings as [BBS17], except we predict 300 emojis classes in instead of 20. We use temporal information as an input to the classifier in addition to the tweet. The date is encoded as a vector of three dimensions, where the first dimension is the month (1-12), the second dimension is the day of the week (1-7), and the last dimension is the local hour (1-24) when the tweet was posted. In the following we describe our classifier architecture with two variants to fuse temporal information with text.

## 4.1 Emoji Prediction Model

We start from the state-of-the-art emoji prediction classifier [BBS17], and built two different methods – *early* and *late* temproral signal fusion– to incorporate the date information. The two entry points for fusing temporal information are evaluated in Section 4.2.

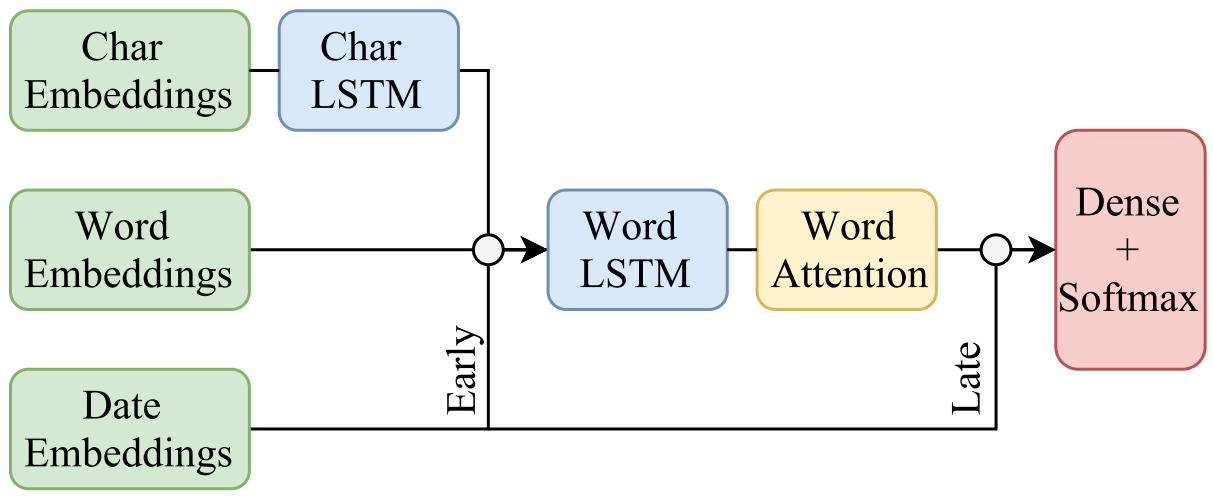

Figure 1: Model architecture where date information is incorporated at an either *early* or late stage.

### 4.1.1 Main Architecture

Inspired by [BBS17], the main architecture begins to extract two different embeddings. The Char B-LSTM takes a sequence of characters and outputs a word embedded vector as in [LLM+15]. The Char B-LSTM output is then concatenated with the word representation as in [BBS17] and passed to the Word LSTM and Word Attention units. We use the attention mechanism introduced in [YYD+16], which can be considered as a weighted average of the output of the Word LSTM, where the weights are learned during training. Finally the fully connected layers and the softmax play the role of the final classifier.

### 4.1.2 Date Module

As previously said the date information is encoded as a vector of three dimensions (month, week day, and hour). For each of these dimensions we create a look up table of vectors of size 10, and vocabulary of 12, 7, and 24 respectively. In this way, we can learn vectors of each month, day of the week and hour, using them for the final classification. These vectors are concatenated all together and incorporated in two different ways in the base system: at an *early* stage or at *late* stage. The *early* stage consists in concatenating this date embeddings to the word representation (char+word embeddings) and pass them to the Word LSTM. The *late* incorporation consists in concatenating the date embeddings with the output vector of the word attention and make the final predictions. We only use one of the methods to include date without combining them.

## 4.2 Emoji Prediction Experiments

We evaluate the automatic prediction of the 300 emojis with three different systems using the method: without date information, using the *early* method to incorporate date information and using the *late* date method.

In Table 5 we report results for the three models using Precision, Recall, Macro F1, Accuracy at 1,3,5,10, and Coverage error. Coverage error (CE) is defined as the average number of labels that have to be in-cluded in the final prediction such that all true labels are predicted.

Table 4: Similarity-matrix results for every combination of two seasons. We show the pairs of emojis with biggest difference among their season specific similarity values.

| | | Spring | Summer | | | Spring | Autumn | | | Spring | Winter | | | Summer | Autumn | | | Summer | Winter | | | Autumn | Winter |
|---|---|---|---|---|---|---|---|---|---|---|---|---|---|---|---|---|---|---|---|---|---|---|---|
| ❗ | 💤 | 0.03 | 0.41 | 🎄 | 🎁 | 0.11 | 0.67 | 🎄 | 🎁 | 0.11 | 0.75 | 🎄 | 🎁 | 0.2 | 0.67 | 🎄 | 🎁 | 0.2 | 0.75 | 😄 | 🔴 | 0.45 | 0.03 |
| 👉 | 💤 | 0.47 | 0.13 | 🏆 | 💉 | 0.31 | 0.7 | ❄️ | 🎁 | 0.12 | 0.56 | 😄 | 🔴 | 0.1 | 0.45 | 💨 | 🏆 | 0.73 | 0.32 | 🎬 | ☝️ | 0.66 | 0.3 |
| 🌟 | 🍑 | 0.23 | 0.55 | ❄️ | 🎁 | 0.12 | 0.5 | 🎉 | 🏆 | 0.7 | 0.27 | 🎁 | 🎊 | 0.89 | 0.55 | 🇺🇸 | 💣 | 0.12 | 0.51 | ❤️ | 💀 | 0.09 | 0.42 |
| 👋 | ✅ | 0.5 | 0.18 | 😄 | 🔴 | 0.07 | 0.45 | 💨 | 🏆 | 0.73 | 0.32 | 🎬 | ☝️ | 0.37 | 0.66 | 🎉 | 💨 | 0.56 | 0.16 | ⚾ | 💀 | 0.05 | 0.37 |
| 🎈 | ❗ | 0.2 | 0.51 | 🎬 | ☝️ | 0.3 | 0.66 | 🎄 | 🍃 | 0.85 | 0.45 | 🎊 | ☝️ | 0.21 | 0.51 | 🏆 | 💉 | 0.23 | 0.61 | 💎 | ❗ | 0.16 | 0.48 |
| 🚨 | 🎵 | 0.14 | 0.45 | 🍴 | ➡️ | 0.46 | 0.08 | 🔥 | 💨 | 0.67 | 0.27 | 🔫 | 🎓 | 0.17 | 0.46 | 🎉 | 🏆 | 0.66 | 0.27 | 🌃 | 🍎 | 0.21 | 0.52 |
| 🍴 | ➡️ | 0.46 | 0.15 | 🔊 | 🚗 | 0.23 | 0.58 | ✨ | 🎁 | 0.3 | 0.66 | 🎧 | 😑 | 0.09 | 0.37 | 🎈 | ❗ | 0.51 | 0.14 | 💀 | ❗ | 0.09 | 0.4 |
| 😁 | 🍷 | 0.29 | 0.59 | ❗ | 💤 | 0.03 | 0.37 | 🎉 | 💨 | 0.53 | 0.16 | 🎤 | 💍 | 0.15 | 0.43 | ❄️ | 🎄 | 0.34 | 0.69 | 😂 | 👏 | 0.58 | 0.28 |
| 😃 | ⚡ | 0.17 | 0.46 | 🎧 | 🚗 | 0.27 | 0.59 | 👑 | 🐝 | 0.66 | 0.3 | 🎧 | 🚗 | 0.23 | 0.59 | 💚 | 🎁 | 0.26 | 0.59 | ☺️ | 💀 | 0.17 | 0.47 |
| ❗ | 😴 | 0.07 | 0.36 | 💄 | 👙 | 0.28 | 0.57 | 🎄 | 🎀 | 0.21 | 0.53 | 🌟 | 🍑 | 0.55 | 0.22 | 💎 | ❗ | 0.15 | 0.48 | ✨ | 🗽 | 0.18 | 0.48 |
| 💛 | 🏈 | 0.48 | 0.76 | 👑 | 🐝 | 0.66 | 0.34 | ❄️ | 🎄 | 0.37 | 0.69 | 😂 | 💔 | 0.49 | 0.2 | 😛 | 😃 | 0.37 | 0.69 | 🚨 | 🎵 | 0.4 | 0.11 |
| 😏 | 📚 | 0.35 | 0.38 | 🎄 | 🍃 | 0.85 | 0.53 | 💚 | 🎁 | 0.27 | 0.59 | 🔊 | 🚗 | 0.29 | 0.58 | 🔑 | 📍 | 0.67 | 0.32 | ☕ | 💀 | 0.06 | 0.35 |
| 🎉 | 💎 | 0.29 | 0.31 | 🎊 | ☝️ | 0.23 | 0.51 | 💋 | 🎄 | 0.16 | 0.48 | 🏆 | 💉 | 0.23 | 0.7 | ✨ | 🎁 | 0.34 | 0.66 | 🔥 | 💨 | 0.55 | 0.27 |
| 🍸 | 🏃 | 0.21 | 0.23 | 💰 | 🎊 | 0.15 | 0.43 | 💥 | 💨 | 0.69 | 0.34 | 💅 | 😚 | 0.42 | 0.38 | 🚨 | 🎵 | 0.45 | 0.11 | 🇺🇸 | 💣 | 0.22 | 0.51 |

Table 5: Results for the three models: whitout date (W/O), *Early* date fudion, and *Late* date fusion. Precision, Recall, F1, accuracy at 1, 5, 10, and Coverage Error.

| | **P** | **R** | **F1** | **a@1** | **a@5** | **a@10** | **CE** |
|---|---|---|---|---|---|---|---|
| **W/O** | 21.97 | 23.22 | 21.89 | 23.13 | 38.22 | 45.70 | 44.29 |
| **Early** | **22.10** | **23.43** | **22.06** | **23.33** | **38.55** | **46.29** | **42.59** |
| **Late** | 21.83 | 23.00 | 21.63 | 22.91 | 37.85 | 45.62 | 43.91 |

Table 6: Single emoji F1 when using Early date information, compared to the model without date information (i.e. W/O). We show the emojis with the biggest improvement. The *Late* fusion model variant is not incorporated as it produces worst performances (see Table 5).

| **Emoji** | 💈 | 🖤 | 💪 | 🍀 | 🍓 | ❤️ | 🌙 | 💨 | ✈️ | ☁️ |
|---|---|---|---|---|---|---|---|---|---|---|
| **W/O** | 0.54 | 0.4 | 0.15 | 0.34 | 0.40 | 0.4 | 0.22 | 0.19 | 0.30 | 0.31 |
| **Early** | 0.63 | 0.11 | 0.22 | 0.41 | 0.47 | 0.10 | 0.27 | 0.24 | 0.35 | 0.36 |
| **Emoji** | ⚠️ | 💅 | 🏐 | 🏁 | 🙋 | 👭 | 🌆 | 👁️ | 🏖️ | 🎥 |
| **W/O** | 0.44 | 0.30 | 0.36 | 0.50 | 0.10 | 0.12 | 0.15 | 0.18 | 0.22 | 0.11 |
| **Early** | 0.49 | 0.34 | 0.40 | 0.54 | 0.14 | 0.16 | 0.19 | 0.22 | 0.26 | 0.15 |

From the results we can see that the best system is the model with *early* incorporation of the date as it outperforms all the other models. The *late* date method is the worst system, even if it seems to create better prediction distributions than the without date system, since the CE is lower. In Table 6 we report the emojis with higher gain in F1, from without date and *early* date, to understand the emojis that depend most on the time information. Among all of the emojis we can see emojis that clearly depend on the month (St. Patrick day, Summer) or the hour (sunrise, moon).

## 5 Conclusions

In the best of our knowledge, this is the first study to investigate if and how temporal information affects the interpretation and prediction of emojis. We studied whether the semantics of emojis change over different seasons, comparing emoji embeddings trained on a corpus of different seasons (Spring, Summer, Autumn, Winter) and show that some emojis are used differently depending on the time of the year, for example 🍀, 🎄, and 🎁. Moreover, we proposed a method to take in account the date information for emoji prediction systems, slightly improving the state-of-the-art. We show that, using the date information, the accuracy of some emojis can be improved. Some of them are cl*early* time dependent (e.g., 🍀, 🌙 ). Others are not directly associated to time but time information helps to predict them ( 💪, 🖤, and ❤️ ).

In the future we plan to study the semantics of emojis over the day (morning/night) or over the week (weekdays/weekend) and improve the date information modules, trying the two methods we proposed together (*early*+late).

## Acknowledgments

Part of this work was done when Francesco B. interned at Snap Inc. Francesco B. and Horacio S. acknowledge support from the TUNER project (TIN2015-65308-C5-5-R, MINECO/FEDER, UE) and the Maria de Maeztu Units of Excellence Programme (MDM-2015-0502).